\documentclass[letterpaper]{article} 
\usepackage{aaai25}  
\usepackage{times}  
\usepackage{helvet}  
\usepackage{courier}  
\usepackage[hyphens]{url}  
\usepackage{graphicx} 
\usepackage{natbib}  
\usepackage{caption} 
\usepackage{algorithm}
\usepackage{algorithmic}

\usepackage{multirow} 
\usepackage{subfigure} 
\usepackage{subcaption} 
\usepackage{mathtools} 
\usepackage{booktabs} 
\usepackage{amsmath} 
\usepackage{amsthm} 
\usepackage{color} 
\usepackage{colortbl} 
\usepackage{bm} 
\usepackage{float} 
\usepackage{amsfonts}       
\usepackage{nicefrac}       
\usepackage{microtype}      
\usepackage{xcolor}         
\usepackage{marvosym}

\usepackage{newfloat}
\usepackage{listings}
\DeclareCaptionStyle{ruled}{labelfont=normalfont,labelsep=colon,strut=off} 
\floatstyle{ruled}
\newfloat{listing}{tb}{lst}{}
\floatname{listing}{Listing}
\title{SGDiff: Scene Graph Guided Diffusion Model for \\Image Collaborative SegCaptioning}
\author{
    Xu Zhang\textsuperscript{\rm 1},
    Jin Yuan\textsuperscript{\rm 1}\thanks{Corresponding author.},
    Hanwang Zhang\textsuperscript{\rm 2},
    Guojin Zhong\textsuperscript{\rm 1},
    Yongsheng Zang\textsuperscript{\rm 1}, \\
    Jiacheng Lin\textsuperscript{\rm 1},
    Zhiyong Li\textsuperscript{\rm 1}\footnotemark[1]
}
\affiliations{
    \textsuperscript{\rm 1}Hunan University, China \\
    \textsuperscript{\rm 2}Nanyang Technological University, Singapore \\
    \{xuzhang1211, yuanjin, gjzhong, pishuai, jcheng\_lin, zhiyong.li\}@hnu.edu.cn, hanwangzhang@ntu.edu.sg\\
    
}

\usepackage{bibentry}

\begin{document}

\maketitle

\begin{abstract}
Controllable image semantic understanding tasks, such as captioning or segmentation, necessitate users to input a prompt (\textit{e.g., text or bounding boxes}) to predict a unique outcome, presenting challenges such as high-cost prompt input or limited information output. This paper introduces a new task ``Image Collaborative Segmentation and Captioning'' (SegCaptioning), which aims to translate a straightforward prompt, like a bounding box around an object, into diverse semantic interpretations represented by (caption, masks) pairs, allowing flexible result selection by users. This task poses significant challenges, including accurately capturing a user's intention from a minimal prompt while simultaneously predicting multiple semantically aligned caption words and masks. Technically, we propose a novel Scene Graph Guided Diffusion Model that leverages structured scene graph features for correlated mask-caption prediction. Initially, we introduce a Prompt-Centric Scene Graph Adaptor to map a user's prompt to a scene graph, effectively capturing his intention. Subsequently, we employ a diffusion process incorporating a Scene Graph Guided Bimodal Transformer to predict correlated caption-mask pairs by uncovering intricate correlations between them. To ensure accurate alignment, we design a Multi-Entities Contrastive Learning loss to explicitly align visual and textual entities by considering inter-modal similarity, resulting in well-aligned caption-mask pairs. Extensive experiments conducted on two datasets demonstrate that SGDiff achieves superior performance in SegCaptioning, yielding promising results for both captioning and segmentation tasks with minimal prompt input.
\end{abstract}

\section{Introduction}
\label{sec:intro}

\begin{figure}[ht!]
	\centering	\includegraphics[width=1\linewidth]{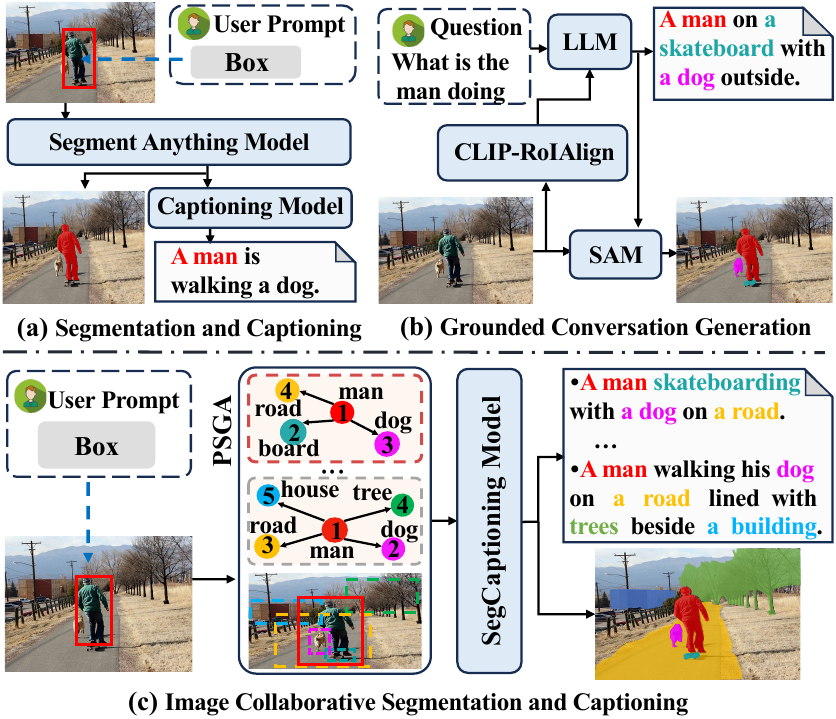}
	\caption{An example to compare different controllable image semantic understanding tasks, where our method uses a box to simultaneously predict multiple aligned bimodal results.}
	\label{fig:framework_comparison}
\end{figure}

The rapid advancement of deep learning has markedly accelerated the evolution of image semantic understanding~\cite{xia2020dilated,xia2020multi}, targeting the extraction of vital semantic information from images through visual or textual representations  \cite{li2017scene,tang2020unbiased}. Classic image semantic understanding tasks encompass vision-based object detection \cite{dai2021dynamic}, image segmentation \cite{xie2021segformer,cheng2022masked}, and image captioning \cite{li2022diffusion,luo2023semantic} etc., where a given image is translated to semantic representations by using boxes, masks, and captions, respectively.

Recently, in pursuit of implementing controllable image semantic expression, researchers have explored enabling users to input prompts to articulate their demands, aiding models in accurately capturing the desired semantic expressions. For instance, referring image segmentation models \cite{wang2022cris} attempt to generate a referring mask based on a sentence. Similarly, controllable image captioning models \cite{cornia2019show} can translate an image into a specific caption according to ordered bounding boxes. Building on the progress of large models, \cite{kirillov2023segment,radford2021learning}, advanced approaches have attempted to generate bimodal outcomes with richer semantic representations guided by prompts. For example, some methods predict a mask and a caption sequentially based on a bounding box \cite{wang2023caption,huang2024segment} (see Figure \ref{fig:framework_comparison}a), while others generate a caption along with corresponding masks guided by a question \cite{rasheed2023glamm} (see Figure \ref{fig:framework_comparison}b). While controllable input facilitates the accurate expression of user needs and encourages models to generate precise results, it also introduces several challenges. Firstly, generating prompt inputs (e.g., a sentence or multiple boxes) is often costly or time-consuming, significantly burdening users. Secondly, existing image semantic understanding methods typically support only unique result generation, which may not align with a user's requirements. Thirdly, although the sequential generation of bimodal representations in Figures \ref{fig:framework_comparison}a and \ref{fig:framework_comparison}b offers richer semantic representations, it also suffers from severe error propagation.

To address these issues, we speculate that it may be feasible for users to input a simple prompt to generate multiple bimodal representations, and 
propose a novel task titled ``Image Collaborative Segmentation and Captioning (SegCaptioning)''. This task concurrently generates intricate bimodal representations for an image, guided by a user's prompt (see Figure \ref{fig:framework_comparison}c). Unlike methods that sequentially predicting bimodal outcomes, SegCaptioning produces coherent bimodal results (a caption alongside related masks) in parallel, leveraging their intricate correlations to mitigate error propagation and enhance performance. Moreover, our task simplifies user input requirements by only necessitating a basic prompt, such as a bounding box to denote the primary object, enabling the system to automatically infer various potential bimodal outcomes, thereby reducing input costs as well as enriching the semantics of outcomes. However, this multifaceted functionality presents notable technological challenges, such as accurately translating a simple prompt into precise intention representations and effectively exploring and aligning correlated bimodal representations.

Towards this end, this paper introduces a pioneering ``Scene Graph Guided Diffusion Model'' (SGDiff), which harnesses structured scene graphs to facilitate intention mapping from a basic prompt to multiple intention representation subgraphs. Additionally, it employs a diffusion process integrated with a bimodal transformer to deeply explore the intricate correlations between bimodal representations. Specifically, when provided with an image and a user's prompt (such as a bounding box around an object), we introduce a Prompt-centric Scene Graph Adaptor (PSGA) to dynamically generate visual scene graphs. This facilitates the accurate capture of users' intentions while effectively filtering out redundant information for caption-mask prediction. 
Building upon this, we devise a diffusion model featuring a Scene Graph Guided Bimodal Transformer (SGBTrans) to predict caption-mask pairs. The diffusion model initially introduces noise to a caption and subsequently denoises it using SGBTrans, which integrates both caption and scene graph features by exploring the complex correlations between them. 
To ensure precise alignment between correlated captions and masks, we devise a Multi-Entities Contrastive Learning Loss (MECL), which explicitly aligns the generated captions and masks by pulling feature representations of positive words and masks closer while pushing away those of negative ones. The MECL operates on both inter-sample and intra-sample levels, yielding well-aligned bimodal results.
Extensive evaluations conducted on MSCOCO and Flickr30k entities datasets validate the effectiveness of SGDiff, achieving state-of-the-art performance compared to existing methods.
In summary, the main contributions of this work can be outlined as follows:
\begin{itemize}
\item We introduce a novel image semantic understanding task called ``Image Collaborative Segmentation and Captioning (SegCaptioning)''. This task enables users to input a straightforward prompt to predict semantic-aligned caption and mask pairs, empowering users to generate richer semantic representations based on their intentions.
\item We propose a novel SGDiff for this new task, facilitating simultaneous generation of (caption, masks) pairs based on a simple prompt, which is translated into a structured scene graph by our proposed PSGA, effectively capturing users' intentions and aiding our model in accurately generating the desired caption-mask pairs.
\item  We design SGBTrans to delve into the intricate correlations between scene graphs and captions throughout the denoising. Additionally, our MECL explicitly aligns each mask and caption word, facilitating effective instance alignment.
\end{itemize}

\begin{figure*}[ht!]
	\centering
	\includegraphics[width=1\linewidth]{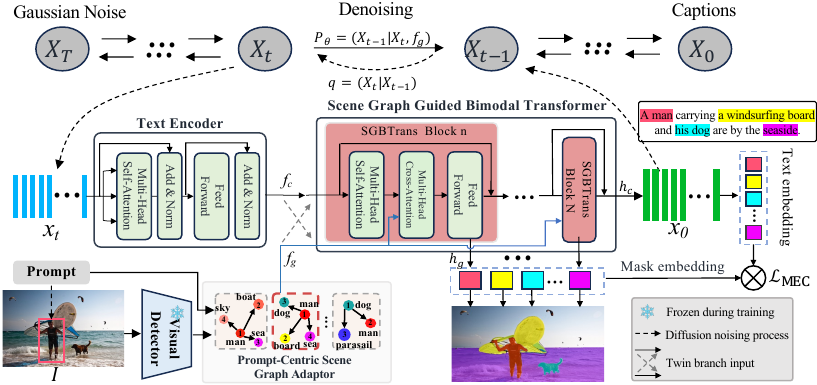}
	\caption{An example to illustrate the training process of our Scene Graph Guided Diffusion Model.}
    \label{fig:networks}
\end{figure*}

\section{Related Work}
\label{sec:related}

\subsection{Image Captioning} \label{sec:related_IC}
Image captioning translates visual content within a scene into a linguistic interpretation, effectively bridging the gap between computer vision and natural language understanding \cite{gan2017stylenet}. Early efforts \cite{yang2019auto,pan2020x} focused on describing visual content from a holistic perspective, resulting in generating limited and fixed captions for any given image. Driven by practical demands, researchers have shifted towards exploring caption generation with local or specific semantics. This shift leds to the development of Controllable Image Captioning (CIC) \cite{deshpande2019fast,wang2024learning}, which utilizes control signals such as clicks \cite{pont2020connecting}, bounding boxes \cite{cornia2019show,chen2021human} and masks \cite{guo2024regiongpt} to generate controllable captions, including semantic controllability \cite{yan2021control}, length controllability \cite{deng2020length}, and sentiment controllability \cite{deshpande2019fast,zhao2020memcap} etc. For instance, \cite{cornia2019show} proposes generating captions based on an ordered sequence of bounding boxes. Meanwhile, \cite{yang2019auto} and \cite{zhong2020comprehensive} employ scene graphs as input to generate image captions, which incorporate both target nodes and relationships to effectively capture users' intentions. 
However, the above autoregressive captioning methods generate text tokens step-by-step, which usually suffers from severe error propagation. Consequently, recent work has explored semi-autoregressive \cite{zhou2021semi} and non-autoregressive diffusion methods \cite{guo2020non, li2022diffusion, luo2023semantic} to enable parallel token output.   
Recently, with the advancement of large models \cite{li2023blip}, CAT \cite{wang2023caption} introduces a universal image captioning approach that integrates various types of prompts (clicks, bounding boxes, scribbles). It utilizes the Segment Anything Model (SAM) \cite{kirillov2023segment} to first extract local visual features conditioned on visual prompts. These visual features are then input into GPT-3 \cite{brown2020language} for caption generation. Instead of adopting the training-free scheme used by CAT, SCA \cite{huang2024segment} introduces a learned module, a ``query-based feature mixer'', to effectively align region-specific visual features with the embedding space of language models for caption generation, yielding improved captioning performance.

\subsection{Image Segmentation} \label{sec:related_IIS}
Traditional image segmentation aims to localize accurate masks concerning semantic objects in an image \cite{kirillov2023segment}. To generate controllable results, researchers have introduced various user prompts \cite{bai2014error}, broadly categorized into two types: visual prompts, like scribbles \cite{wang2018deepigeos}, boxes \cite{zhang2020interactive}, and clicks \cite{chen2021conditional,zhang2024pvpuformer}; and textual prompts \cite{wang2022cris, lin2023brppnet}. For instance, \cite{lin2020interactive} optimizes the use of initial clicks to enhance outcomes, while \cite{zou2024segment} uses scribbles to provide richer and more precise information to capture users' intentions, albeit at a higher labeling cost. Compared to visual prompts, language prompts can increase user burden but more accurately express users' intentions, thereby helping models better predict referring masks \cite{luo2020multi, liu2023gres}. Inspired by the complementarity of various prompts, SAM \cite{kirillov2023segment} first unifies various types of user prompts—including a set of clicks, a region box/mask, or free-form text—to establish a versatile foundational model for segmentation. This offers a flexible prompt interface for users. Additionally, it is the first visual large model for image segmentation.
Building on this foundation, GROUNDHOG \cite{zhang2024groundhog} leverages SAM to obtain region mask proposals through visual prompts and then uses Multimodal Large Language Models (MLLMs) to generate grounded captions based on regional semantic features. GLaMM \cite{rasheed2023glamm} aligns visual regions with text tokens using SAM and GPT-4 \cite{openai2023gpt4}, first generating a caption by querying GPT-4 and then producing corresponding region masks with SAM guided by the caption. Although the bimodal outcomes enrich the content representations, the sequential predictions usually suffer from severe error propagation.

Different from \cite{yang2022diffusion,zou2023generalized,ren2024pixellm,huang2024segment}, 
our approach only requires a simple visual prompt to infer user needs. This prompt guides the generation of multiple bimodal outcomes, enriching the output content. Moreover, our method predicts both masks and captions in parallel, alleviating the error propagation issues associated with the sequential prediction methods.

\section{Methodology}  \label{sec:approach}

\subsection{Overview} \label{approach:problem}

Given an image $\boldsymbol{I} \in \mathbb{R}^{H \times W \times 3}$ with height $H$ and width $W$, , along with a user's prompt $P_o$ such as a box bounding an object $o$, SegCaptioning aims to generate multiple (caption, masks) pairs. Here, we use $S$ to denote a caption centered around $o$, and $M$ to represent several masks related with $S$. To achieve this goal, we design a novel ``Scene Graph Guided Diffusion '' (SGDiff) model tailored for this specific task. As illustrated in Figure \ref{fig:networks}, our proposed model comprises three key components: a textual encoder that extracts textual features for captions, a Prompt-Centric Scene Graph Adaptor (PSGA) that transforms a prompt-guided image to a scene graph for effective intention capturing, and a diffusion model with a Scene Graph Guided Bimodal Transformer (SGBTrans) that predicts caption-mask pairs. Furthermore, our model introduces a MECL to explicitly align masks with their corresponding captions. This ensures that the bimodal results are well-aligned and coherent, enhancing the quality and relevance of the generated content.

\subsection{Prompt-Centric Scene Graph Adaptor} \label{approach:vencoder}
Given an image $\boldsymbol{I}$ along with a caption-mask pair $(S, M)$, the proposed PSGA receives a user's prompt and then transforms $I$ into a scene graph $G_i$ to encapsulate the caption-mask pair. Specifically, our approach first generates a global scene graph $G$ following \cite{zellers2018neural}. Subsequently, it predicts a subgraph $G_i$ within $G$ according to a user's prompt, facilitating the subsequent prediction on $(S, M)$. Concretely, when a user inputs a prompt $P_o$ referring to the center object $o$ in $(S, M)$, PSGA aims to search for a subgraph $G_i$ from $G$ to represent $(S, M)$, which is mathematically expressed as: 
\begin{equation}
\mathop{\arg\max}_{G_i} P(G_i=(N, E)|G,o).
\end{equation}
To implement this, our approach initially generates a coarse subgraph $G_i=(N, E)$, where $N$ and $E$ denote the node and edge sets connected to $o$, respectively. As the numbers of nodes and edges in $G_i$ typically exceed the desired ones in $(S, M)$, our approach further constructs an adaptor to refine $G_i$. Specifically, the adaptor comprises several self-attention blocks and one mapping layer. Initially, we feed the node feature $e_o \in \mathbb{R}^{L\times D}$ of $G_i$ into the self-attention blocks to produce an updated feature $f_o \in \mathbb{R}^{L\times D}$, where $L$ is the number of nodes and $D$ is the feature dimension. Subsequently, the mapping layer predicts node relevance scores $P(G_i) \in \mathbb{R}^{L}$, representing the node relevance concerning $(S, M)$. We train the adaptor using the Binary Cross-Entropy (BCE) loss to minimize $P(G_i)$ and its ground truth $\hat{P}(G_i)$:
\begin{equation}
\mathcal{L}_{\text{adaptor}} = \text{BCE}(P(G_i),\hat{P}(G_i)).
\end{equation}
For each object in $(S, M)$, we compute its Intersection over Union (IOU) with the corresponding node in $G_i$. If the IOU exceeds a threshold, we assign the score in $\hat{P}(G_i)$ as $1$, otherwise $0$. Consequently, the adaptor is able to effectively predict relevant objects in $G_i$ while filtering out irrelevant ones.

Furthermore, the random order of objects in the node features $f_o$ may  not align with the appearance order in the caption $S$, potentially impacting caption generation, as discussed in \cite{cornia2019show}. Hence, we consider measuring the order consistency between the node feature and the caption. To address this, we employ a multi-layer fully connected network to map $f_o$ into a permutation matrix $R \in \mathbb{R}^{L\times K}$, where the $L$ rows represent the $L$ objects in $f_o$ with a random order, and the $K$ columns record the ordered objects from the caption $S$. The element $r_{lk}$ indicates the probability of the $l$-th object in $f_o$ semantically consistent with the $k$-th object in $S$. We design a ranking loss $\mathcal{L_{\text{ranking}}}$ to minimize the distance between the predicted matrix $R$ and the ground truth matrix $\hat{R}$ using a Cross-Entropy (CE) loss:
\begin{equation}
\mathcal{L_{\text{ranking}}} = \text{CE}(R, \hat{R}).
\end{equation}
After filtering out irrelevant nodes, we adjust the node order to match that of the caption, resulting in a refined feature $f'_o$:
\begin{equation}
f'_o = \text{Ranking} (f_o \times {\delta(P(G_i))}),
\label{reranking}
\end{equation}
\begin{equation}
    {\delta(x)}=\left\{\begin{array}{ll}
    x, & \text { if } x \geq \theta \\
    0, & \text { otherwise, }
    \end{array}\right. 
\label{eq:filter}
\end{equation}
where $\theta$ is a threshold to filter irrelevant objects. $\text{Ranking}$($\cdot$) rearranges the row vectors in $f_o$ so that its feature order is consistent with the object order in the caption. On this basis, our approach fuses $f'_o$ and the edge feature $f_e$ following \cite{guo2019aligning}, generating the graph feature $f_g$.

The PSGA is trained using a composite loss $\mathcal{L_{\text{SGadaptor}}} = \mathcal{L_{\text{adaptor}}} + \mathcal{L_{\text{ranking}}}$, offering several advantages: Firstly, it effectively maps a user's prompt to a concise and accurate subgraph consistent with the to-be-predicted caption, retaining only relevant information to facilitate subsequent predictions. Secondly, the generated graph feature maintains a consistent object order with the caption for performance improvement. In the testing phrase, we use the PSGA to generate several related subgraphs to predict multiple (caption, masks) pairs. 

\subsection{Scene Graph Guided Diffusion Model} \label{approach:Diffusion}
Similar to Bit Diffusion \cite{chen2022analog}, our SGDiff adds Gaussian noise into a caption sample $\mathbf{x}$ from timestep $t=0$ to $T$ during the forward process and then performs a reverse denoising process to recover the caption. 
Differently, we introduce a scene graph feature $f_g$ to strengthen the denoising to simultaneously predict a caption with related masks. Specifically, given a noised sample $\mathbf{x}_t$ and $f_g$, SGDiff iteratively generates noise-free text descriptions $x_0$ by progressively denoising $x_t$, and each denoising transition $x_{t} \rightarrow x_{t-1}$ is parametrized by a model $p_{\theta}(x_{t-1}\mid x_{t}, f_{g})$:
\begin{small}
\begin{equation}
p_{\theta}(\mathbf{x}_{0:T}):=p(\mathbf{x}_T)\prod_{t=1}^Tp_{\theta}(\mathbf{x}_{t-1}|\mathbf{x}_t, f_g).
\label{eq:denoising}
\end{equation}
\end{small}
We design a scene graph guided bimodal transformer to implement Eq. \eqref{eq:denoising} with a  training loss:
\begin{equation}
\mathcal{L}_{\text{bit}} = {\mathbb{E}_{t \sim \mathcal{U}(0,T),\epsilon  \sim \mathcal{N}(0,\textit{I})}}{\left\| {f({x_t}, t, f_g) - {x_0}} \right\|^2},
\label{eq:diff_forward}
\end{equation}
where $t\sim \mathcal{U}(0, T)$, $\bm \epsilon \sim \mathcal{N}(\bm 0, \bm I)$ are continuous variables.

\textbf{Scene Graph Guided Bimodal Transformer.} 
At each time step $t$, SGBTrans first receives a noised caption feature $f_c$ and the graph feature $f_g$, and then predicts bimodal features including a denoised text feature and a visual feature. Specifically, SGBTrans comprises $B$ blocks, each containing a multi-head self-attention layer, a multi-head cross-attention layer, and a feed-forward layer. Unlike uni-modal generation transformers, our task involves predicting a caption with masks. Therefore, we devise a cross-training scheme for bimodal generation. To elaborate, for caption generation, we utilize $f_c$ as a query and $f_g$ as value and key into the blocks:
\begin{small}
\begin{equation}
\begin{aligned}
&f_{c}^{'} = {\bf{MultiHead_s}}(f_{c},f_{c},f_{c}),\\
&u_{c} = {\bf{FFN}}(f_{c}^{'} + {\bf{MultiHead_c}}(f_{c}^{'},f_g,f_g)), \\
&h_c = {\bf{\phi}_c}(u_{c} + f_{c}).  \label{caption}
\end{aligned}
\end{equation}
\end{small}
where ${\bf{MultiHead_s}}$ and ${\bf{MultiHead_c}}$ represent the multi-head self-attention and cross-attention layers, respectively, ${\bf{FFN}}$ denotes the feed-forward layer, and ${\bf{\phi}}( \cdot )$ indicates the mlp layer. The resulting $h_c$ represents a refined caption feature, which is then either passed to the subsequent block or utilized for caption prediction. Compared to the global visual feature, the scene graph feature provides a more precise visual description, aligning better with the caption to be predicted and thereby enhancing caption generation. Regarding mask generation, we input $f_g$ as a query and $f_c$ as both value and key into the blocks as follows:
\begin{small}
\begin{equation}
\begin{aligned}
&f_{g}^{'} = {\bf{MultiHead_s}}(f_{g},f_{g},f_{g}),\\
&u_{g} = {\bf{FFN}}(f_{g}^{'} + {\bf{MultiHead_c}}(f_{g}^{'},f_c,f_c)), \\
&h_g = {\bf{\phi}_g}(u_{g} + f_{g}).
\label{mask}
\end{aligned}
\end{equation}
\end{small}
Guided by the caption feature, $h_g$ could accurately offer visual features for mask prediction.
Unlike  \cite{yang2022diffusion,sun2023sgdiff}, 
Our method distinguishes itself by using PSGA to convert a visual prompt into object-centered scene graphs, guiding the generation of multiple bimodal outputs.

\textbf{Caption and Mask Generation.} We utilize the denoised caption feature $h_c$ to predict the probability distribution for each word in the caption. Further details can be found in \cite{luo2023semantic}, where the loss  $\mathcal{L_{\text{caption}}}$ contains a diffusion loss in Eq. \eqref{eq:diff_forward} and a cross-entropy loss $\mathcal{L_{\text{CE}}}$ with equal weights. For mask generation, we adopt the Mask2Former \cite{cheng2022masked}, which takes the global visual feature $f_v$ extracted by Swin-Transformer \cite{liu2021swin} and the graph $h_g$ in Eq. \eqref{mask} as input to produce class-agnostic binary region masks. The mask loss $L_{\text{Mask}}$ employs both Dice loss and BCE loss.

\subsection{Multi-Entities Contrastive Learning} \label{approach:contrastive}

We propose a MECL to explicitly explore cross-modal representation correlations between visual entities and caption words. Our motivation is that matching mask-word pairs should have similar contextual representations. Given a predicted mask-caption pair $(M, S)$, it's important to note that the relationship between masks and words is not strictly one-to-one. For instance, the word \textit{dog} may refer to multiple mask entities, and conversely, a mask may contain objects indicated by multiple words. Therefore, the $i$-th mask embedding $\mathbf{m}_i$ is expected to be close to the matched word embedding sets ${\mathbf{s}_i }^+$ while being distant from other word embeddings $\mathbf{s}_j$. In other words, the mutual information between mask entities and words that are matched should be maximized. Since directly optimizing mutual information is infeasible, we design an intra-sample contrastive loss $\mathcal{L}_{\text{intra}}$ inspired by InfoNCE \cite{oord2018representation}, representing a lower bound of mutual information. This loss is formulated as follows:
\begin{small}
\begin{equation}
\begin{split}
\mathcal{L}_{\text{intra}} = &\!- \sum_{i=1}^{N} \frac{1}{\vert \{ \mathbf{s}_i \}^+ \vert } \sum_{\mathbf{s}_i \in  \{ \mathbf{s}_i \}^+ } \!\!\!\!\!\log \frac{\exp(\mathbf{m}_i^\top  \mathbf{s}_i / \tau)}{\sum\limits_{j \neq i} \exp(\mathbf{m}_i^\top \mathbf{s}_j/\tau) } \\
&- \!\!\sum_{i=1}^{L} \frac{1}{\vert \{ \mathbf{m}_i \}^+ \vert} \sum_{\mathbf{m}_i \in \{ \mathbf{m}_i \}^+} \!\!\!\!\!\log \frac{\exp( \mathbf{s}_i^\top  \mathbf{m}_i / \tau)}{\sum\limits_{j \neq i} \exp(\mathbf{s}_i^\top  \mathbf{m}_j /\tau) } ,
\end{split}
\end{equation}
\end{small}
where $\tau$ is a learnable parameter, $N$ and $L$ are the number of mask and word tokens. Likewise, we impose that the $i$-th word embedding $\mathbf{s}_i$ is similar to the matched mask embedding sets $\{\mathbf{m}_i \}^+$ and dissimilar to the irrelevant $\mathbf{m}_j$.

Aligning mask and word embeddings solely at the intra-sample level may be insufficient as it presents challenges in exploring the discriminability of visual and textual representations across different samples. Consequently, we propose inter-sample contrastive learning to capture the discriminative information of multiple mask-caption pairs. Specifically, we define the global matching score for each mask-caption pair $g(\mathbf{m},\mathbf{s})$ as the weighted average of their mask-word pair scores. This allows us to consider the relationships between masks and words across multiple samples:
\begin{small}
\begin{align}
    & g(\mathbf{m}, \mathbf{s}) = \frac{1}{L}\sum_{j=1}^L\sum_{i=1}^N \rho(\mathbf{m}, \mathbf{s})_{i,j}{\cdot} \langle \mathbf{m}_i, \mathbf{s}_j \rangle,\label{eq:similarity}
\end{align}
\end{small}
where $\langle \cdot, \cdot \rangle$ is the dot product of two vectors and $\rho(\mathbf{m},\mathbf{s})_{i,j}=\frac{\exp \langle \mathbf{m}_i,\mathbf{s}_j \rangle}{\sum^N_{i^{\prime}=1}\exp \langle \mathbf{m}_{i^{\prime}},\mathbf{s}_j \rangle}$. Then the inter-sample contrastive loss $\mathcal{L}_{\text{inter}}$ can be computed by the random pairs between mask and caption in the same minibatch $B$ as follows:
\begin{small}
\begin{equation}
\begin{split}
    \mathcal{L}_{\text{inter}} = &-\frac{1}{B}\sum_{i=1}^B\log \frac{\exp(g(\mathbf{m}^i, \mathbf{s}^i) /\tau) }{\sum_{j\neq i} \exp(g(\mathbf{m}^i, \mathbf{s}^j) /\tau)}\\
    &-\frac{1}{B}\sum_{i=1}^B\log \frac{\exp(g(\mathbf{s}^i, \mathbf{m}^i) /\tau) }{\sum_{j \neq i} \exp(g(\mathbf{s}^i, \mathbf{m}^j) /\tau)},
\end{split}
\end{equation}
\end{small}
where the matched mask-caption pairs with the same index will be pulled closer in the shared semantic representation space and $\mathcal{L}_{\text{inter}}$ can be viewed as an optimization of the global mutual information between visual entities and caption. Toward this objective, the MECL can be expressed as $\mathcal{L}_{\text{MEC}} = \mathcal{L}_{\text{inter}} + \mathcal{L}_{\text{intra}}$, designed to complement representation learning and further facilitate joint multi-modal learning within the diffusion process.
Finally, it's worth noting that we train the entire SGDiff model using either $\mathcal{L}_\text{Caption}$ or $\mathcal{L}_{\text{Mask}}$, alongside the proposed $\mathcal{L}_{\text{MEC}}$ and $\mathcal{L}_{\text{SGadaptor}}$ with balancing weights ${\lambda}_{\{1,2\}}$. 
The total loss is calculated as:
\begin{equation}
  \mathcal{L}_{\mathrm{total}} = \mathcal{L_{\text{Caption}}} + \mathcal{L_{\text{Mask}}} + {\lambda}_1 \mathcal{L_{\text{SGadaptor}}} + {\lambda}_2 \mathcal{L_{\text{MEC}}}.
\label{eq:total_loss}
\end{equation}

\section{Experiments}  \label{experiments}

\begin{figure*}[t!]
	\centering
\includegraphics[width=1\linewidth]{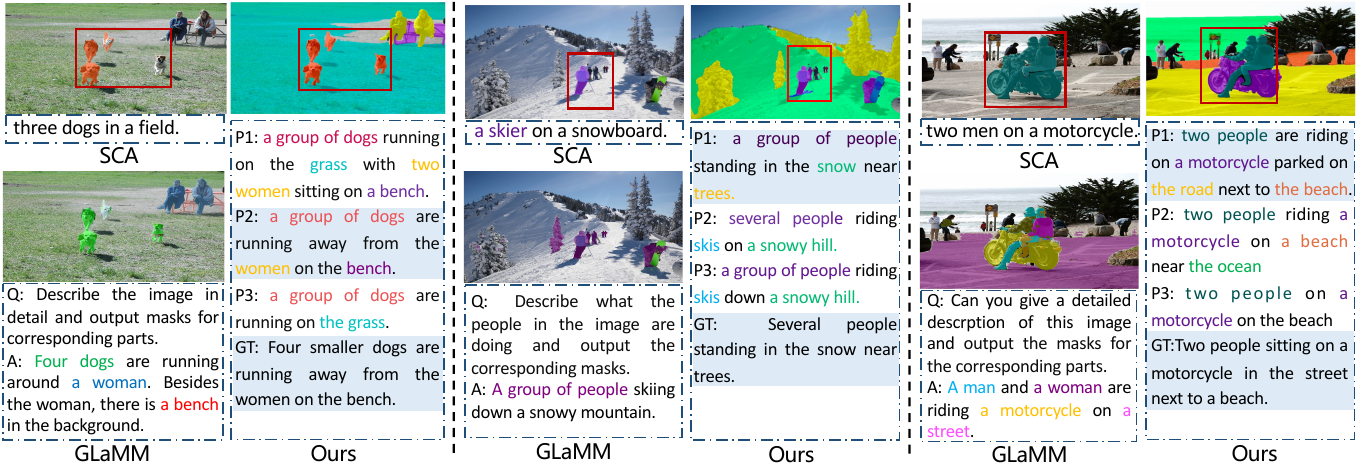}
	\caption{Quantitative results of our SGDiff, SCA, and GLaMM on MSCOCO, where color indicates the matching relationship between masks and words.}
	\label{fig:results}
\end{figure*}

\subsection{Datasets and Metrics}
\textbf{MSCOCO} \cite{lin2014coco} comprises $123,287$ images, each annotated with five captions. For each caption, we utilize all the nouns to identify the corresponding masks provided in \cite{lin2014coco}, resulting in a (caption, masks) pair for each image across $133$ categories. 
We adhere to the Karpathy split \cite{karpathy2015deep} to allocate $113,287$ images for training, $5,000$ for validation, and $5,000$ for testing.

\textbf{Flickr30k Entities} \cite{plummer2015flickr30k} is an extension of Flickr30k~\cite{young2014image},  consisting of $31,000$ images with five captions for each one. For each caption, noun entities are linked to one or more boxes. We employ SAM \cite{kirillov2023segment} to approximately convert each box into a binary mask, thereby matching each caption with several masks. 
We follow the split suggested by Karpathy \cite{cornia2019show}, designating $29,000$ images for training, and $1,000$ for validation and testing, respectively.

We evaluate our model on three aspects: caption quality, mask quality, and caption-mask matching. Caption quality is assessed using Blue-4, METEOR, SPICE and CIDEr metrics. Mask quality is evaluated with Mask IoU. Caption-mask matching is measured by calculating the mAP of mask classification for objects in the caption. Since our method supports the generation of multiple results, we select the top 5 results and calculate the highest captioning score as the final result for comparison.

\subsection{Implementation Details} 

To train our model, we employ large-scale jittering on each input image with a size of $1024^2$. We approximately use the first noun in each caption as the central object, and its corresponding bounding box as a user's prompt. Our training consists of two stages. First, we train the model for captioning by deleting $\mathcal{L_{\text{Mask}}}$ and $\mathcal{L_{\text{MEC}}}$ in Eq. \eqref{eq:total_loss}, where the SGDiff network is optimized using the Adam optimizer with a learning rate of $0.0001$ and a weight decay of $0.05$ on two A6000 GPUs. The optimization process spans $60$ epochs with a batch size of $16$. Second, we jointly train the mask and caption generation with the complete loss. The optimization process spans $60$ epochs with a batch size of $16$.

\begin{table}[t!]
    \centering
    \scriptsize
    \renewcommand\arraystretch{1.0}
    \setlength\tabcolsep{0.7mm}
    \begin{tabular}{l|cc|cc|cc}
    \toprule
    Model & SEG & CAP & SPICE & CIDEr & mIoU & mAP \\ \midrule
    Sub-GC $_{\mathrm{ECCV20}}$ & - & $\checkmark$ &  20.2   & 115.3  & - & - \\
    SATIC $_{\mathrm{ICCV21}}$  & - & $\checkmark$ & 22.3   & 127.2  & - & -   \\
    X-Transformer  $_{\mathrm{CVPR20}}$ & - & $\checkmark$ & 23.4   & 132.8  & - & -   \\
    CMAL $_{\mathrm{IJCAI20}}$  & - & $\checkmark$ &  21.8   & 124.0  & - & - \\
    SCD-Net $_{\mathrm{CVPR23}}$  & - & $\checkmark$ & 23.0   & 131.6  & - & -  \\
    \hline
    OpenSeg $_{\mathrm{ECCV22}}$ & $\checkmark$ & - & -   & - & 38.1  & -   \\
    OpenSeeD $_{\mathrm{ICCV23}}$ & $\checkmark$ & - & -   & -  & 64.0 & \textbf{47.6}   \\
    ODISE $_{\mathrm{CVPR23}}$ & $\checkmark$ & - & -   & - & 65.2  & 46.0   \\
    \hline
    X-Decoder $_{\mathrm{CVPR23}}$ & $\checkmark$ & $\checkmark$ & -   & 122.3  & 62.4 & 41.3   \\
    CAT $_{\mathrm{arXiv23}}$ & $\checkmark$ & $\checkmark$ & 17.9   & 90.1 & 24.9  & 8.2   \\
    SCA $_{\mathrm{CVPR24}}$ & $\checkmark$ & $\checkmark$ & 18.8   & 85.0 & 32.4  & 11.3   \\
    GLaMM $_{\mathrm{CVPR24}}$  & $\checkmark$ & $\checkmark$ & 23.3   & 130.9  & 36.2 & 14.8   \\
    SGDiff (ours) & $\checkmark$ & $\checkmark$ & \textbf{26.1}   & \textbf{137.4} & \textbf{66.3}  & 47.2  \\ 
    \bottomrule
    \end{tabular}
        
    \caption{Performance comparison with several advanced methods on MSCOCO. 
    SEG and CAP denote Segmentation and Captioning, respectively.}
    \label{results_coco}
\end{table}

\begin{table}[t!]
    \centering
    \scriptsize
    \renewcommand\arraystretch{1.0}
    \setlength\tabcolsep{0.7mm}
    \begin{tabular}{l|cccc}
    \toprule
    Model & Bleu-4 & METEOR  & SPICE & CIDEr  \\ \midrule
    SCT $_{\mathrm{CVPR19}}$  & 12.5 & 16.8 &  23.5  & 84.0   \\
    Sub-GC $_{\mathrm{ECCV20}}$  & 11.2 & 15.9 &  22.2   & 73.7   \\
    VSR $_{\mathrm{CVPR21}}$  & 10.7 & 18.0 & 21.9   & 97.5   \\
    ComPro $_{\mathrm{IJCV24}}$  & 11.9 & 17.3 & 23.9   & 89.4   \\
    CAT $_{\mathrm{arXiv23}}$   & 30.2 & 22.7  & 15.6   & 63.4   \\
    SCA $_{\mathrm{CVPR24}}$   & 20.8 & 17.9  & 13.4   & 35.4   \\
    GLaMM $_{\mathrm{CVPR24}}$  & 35.4 & 26.4   & 18.8   & 95.3   \\
    GROUNDHOG $_{\mathrm{CVPR24}}$   & 36.7   & 26.5 & 20.4 & 91.3   \\
    SGDiff (ours)   & \textbf{38.2} & \textbf{27.4}  & \textbf{24.5} & \textbf{98.6}   \\ 
    \bottomrule
    \end{tabular}
    
    \caption{Performance comparison with several advanced methods on Flickr30k Entities.}
    \label{results_flickr30k}
\end{table}

\subsection{Comparison with State-of-the-Art Methods}

This is the first study to support generating multiple caption-mask pairs. We benchmark SGDiff against advanced methods for generating masks, captions, or both, as shown in Tables \ref{results_coco} and \ref{results_flickr30k}. 
Compared to captioning methods like Sub-GC, SCA, and CAT, SGDiff outperforms them due to its precise capture of user intentions through the PSGA and its effective prediction capabilities enabled by the scene graph-guided diffusion process. 
For segmentation, we compare SGDiff with several open-vocabulary approaches (OpenSeg, OpenSeeD, and ODISE) on the MSCOCO dataset. Our SGDiff excels due to its thorough exploration and strict alignment between masks and words. It also surpasses large models like CAT, SCA, and GLaMM in both tasks. Notably, SGDiff facilitates collaborative prediction in both mask and caption generation, which significantly mitigates error propagation and boosts performance. 
Figure \ref{fig:results} shows quantitative results comparing SGDiff, SAC, and GLaMM on MSCOCO. SAC is limited to generating a text description with masks confined to a bounding box and lacks the capability to capture semantic and visual representations outside of it, resulting in a marked performance decline (as shown in Table \ref{results_coco}). GLaMM, which requires users to type a sentence, shows relatively low segmentation performance since its SAM struggles to accurately predict masks guided by text. 
In contrast, our SGDiff supports the generation of multiple bimodal results, offering diverse yet precise outcomes guided by a simple prompt, thus reducing user burden and enhancing overall performance.

\subsection{Analysis of PSGA}

\begin{table}[t!]
    \centering
    \scriptsize
    \begin{tabular}{cc|cc|c|c}
    \hline 
    Filtering    & Ranking    & SPICE & CIDEr & mIOU & mAP \\ \midrule
    -            & -            & 24.2     & 127.4   & 62.3     & 43.2      \\
    $\checkmark$ & -            & 24.9   & 133.7 & 64.2  & 45.4      \\
    -            & $\checkmark$            & 25.4   & 135.8 & 65.5  & 46.3      \\ 
    $\checkmark$ & $\checkmark$            & \textbf{26.1}   & \textbf{137.4} & \textbf{66.3}  & \textbf{47.2}      \\ 
    \hline
    \end{tabular}
    
    \caption{Performance comparison of PSGA by using different components.}
    \label{PSGA}
\end{table}

\begin{table}[t!]
    \centering
    \scriptsize
    \renewcommand\arraystretch{1.0}
    \begin{tabular}{c|cc|c|c}
    \hline
    $\lambda_1$ & SPICE & CIDEr & mIOU & mAP \\ \midrule
    0       & 24.2     & 127.4   & 62.3     & 43.2       \\
    1       & 24.9     & 135.7   & 65.7     & 46.0       \\
    2       & \textbf{26.1}   & 137.4 & \textbf{66.3}  & \textbf{47.2}      \\ 
    5       & 25.8     & \textbf{137.6}   & 65.9     & 46.6      \\
    10      & 25.3   & 136.5 & 65.4  & 45.8      \\ \hline
    \end{tabular}
    
    \caption{Performance change of the loss $\mathcal{L_{\text{SGadaptor}}}$ with different weights on MSCOCO.}
    \label{PSGA2}
\end{table}

Our PSGA operates on a coarse subgraph by performing node filtering (see Eq. \eqref{eq:filter}) and ranking (see Eq. \eqref{reranking}). This experiment confirms the effectiveness of these two components, as illustrated in Table \ref{PSGA}. Node filtering removes irrelevant objects, while node ranking ensures a consistent order between graph nodes and caption words, both significantly improving performance.  
Table \ref{PSGA2} details the performance changes with respect to different weights $\lambda_1$ in Eq. \eqref{eq:total_loss} under $\lambda_2=1$. When $\lambda_1=0$, ignoring PSGA training, performance is lowest. Increasing $\lambda_1$ from 0 to 2 consistently improves performance, demonstrating the effectiveness of PSGA. 
However, larger $\lambda_1$ values lead to performance declines, as an excessively large value may diminish the utility of other losses.

\subsection{Analysis of SGBTrans}

\begin{table}[t!]
    \centering
    \scriptsize
    \renewcommand\arraystretch{1.0}
    \setlength\tabcolsep{1mm}
    \begin{tabular}{c|cc|c|c}
    \hline
    SGB-Trans  & SPICE & CIDEr & mIOU & mAP \\ \midrule
    without cross-attention      & 24.5     & 133.7   & 64.5     & 46.0      \\
    with cross-attention         & \textbf{26.1}   & \textbf{137.4} & \textbf{66.3}  & \textbf{47.2}      \\ \hline
    \end{tabular}
    
    \caption{Performance comparison by using SGBTrans with or without cross-attention on MSCOCO.}
    \label{ablation_sct}
\end{table}

\begin{table}[t!]
    \centering
    \scriptsize
    \renewcommand\arraystretch{1.0}
    \begin{tabular}{cc|cc|c|c}
    \hline
    $\mathcal{L_{\text{Mask}}}$    & $\mathcal{L_{\text{Caption}}}$    & SPICE & CIDEr & mIOU & mAP \\ \midrule
    -            & $\checkmark$            & 25.2     & 135.8   & -     & -      \\
    $\checkmark$ & -            & -     & -   & 63.7     & 44.5      \\
    $\checkmark$ & $\checkmark$    & \textbf{26.1}   & \textbf{137.4} & \textbf{66.3}  & \textbf{47.2}      \\
    \hline
    \end{tabular}
    
    \caption{Performance comparison of our SGDiff with individual tasks on MSCOCO.}
    \label{results_single}
\end{table}

\begin{table}[t!]
    \centering
    \scriptsize
    \renewcommand\arraystretch{1.0}
    \begin{tabular}{cc|cc|c|c}
    \hline 
    $L_{intra}$    & $L_{inter}$    & SPICE & CIDEr & mIOU & mAP \\ \midrule
    -            & -            & 24.8     & 134.8   & 64.2     & 42.7      \\
    $\checkmark$ & -            & 25.4     & 136.4   & 65.6     & 46.2      \\
    -            & $\checkmark$ & 25.6     & 136.7   & 65.3     & 46.0      \\ 
    $\checkmark$ & $\checkmark$ & \textbf{26.1}   & \textbf{137.4} & \textbf{66.3}  & \textbf{47.2}      \\ 
    \hline
    \end{tabular}
    
    \caption{Performance comparison of the $\mathcal{L_{\text{MEC}}}$ with different components on MSCOCO.}
    \label{ablation_loss}
\end{table}

\begin{table}[t!]
    \centering
    \scriptsize
    \renewcommand\arraystretch{1.0}
    \begin{tabular}{c|cc|c|c}
    \hline
    $\lambda_2$ & SPICE & CIDEr & mIOU & mAP \\ \midrule
    0       & 24.8     & 134.8   & 64.2     & 42.7       \\
    1       & \textbf{26.1}     & \textbf{137.4}   & \textbf{66.3}     & \textbf{47.2}       \\
    2       & 25.6     & 137.1   & 66.1     & 47.0      \\ 
    5       & 25.4     & 136.8   & 65.8     & 46.4      \\
    10      & 25.0     & 136.2   & 65.5     & 46.1      \\ \hline
    \end{tabular}
    
    \caption{Performance comparison of the $\mathcal{L_{\text{MEC}}}$ with different weights on MSCOCO.}
    \label{ablation_lossweight}
\end{table}

Our SGBTrans takes a noised caption feature and a scene graph feature, facilitating deep interaction for caption-mask prediction. This experiment compares the performance of SGBTrans with and without the cross-attention layer (see Eq. \eqref{caption} and Eq. \eqref{mask}), as shown in Table \ref{ablation_sct}. Incorporating the cross-attention operation between caption and graph features significantly enhances performance since the cross-modal interaction compensates for the deficiencies of each feature, thereby guiding the generation of accurate captions and masks.

\subsection{Analysis of Collaborative SegCaptioning}

This experiment verifies the effectiveness of collaborative segmentation and captioning, as shown  in Table \ref{results_single}. Our SGDiff employs $\mathcal{L_{\text{Mask}}}$, $\mathcal{L_{\text{Caption}}}$, or both.
It is evident that collaborative training of both tasks achieves superior performance by effectively exploring the intricate correlations between captions and masks.
Furthermore, our approach introduces the Multi-Entities Contrastive Loss ($\mathcal{L}_{\text{MEC}}$), comprising $\mathcal{L}_{\text{intra}}$ and $\mathcal{L}_{\text{inter}}$ for mask-caption alignment, as shown in Table \ref{ablation_loss}. Both components enhance performance, particularly mAP, by aligning masks and words at intra-sample and inter-sample levels.  
Table \ref{ablation_lossweight} further demonstrates the overall effectiveness of $\mathcal{L}_{\text{MEC}}$ with different weights in Eq. \eqref{eq:total_loss}, where the optimal performance is achieved when $\lambda_2=1$.
Increasing $\lambda_2$ beyond this point results in a performance decline, as it tends to diminish the utilities of the other losses.

\section{Conclusion}  \label{conclusion}
This paper introduces a novel ``Scene Graph Guided Diffusion Model'' for the newly defined task of ``Image Collaborative SegCaptioning''. It represents the first exploration of this topic, providing valuable insights and guidance for future research. Technically, we develop an innovative PSGA that automatically translates a simple prompt into multiple related scene graphs, effectively inferring a user's intention. Building on this, we establish an effective diffusion pipeline that integrates a SGBTrans to explore the complex interplay between scene graph and text inputs, as well as a MECL to explicitly align each mask with its corresponding caption word, producing diverse yet precise caption-mask pairs that meet user requirements. Extensive experiments on two benchmark datasets demonstrate the effectiveness of our approach.

\section{Acknowledgments}
The authors would like to thank the reviewers for their comments that help improve the manuscript. 
This research is supported by the National Natural Science Foundation of China (Grants No.~62272157, U21A20518, and U23A20341), the RIE2025 Industry Alignment Fund – Industry Collaboration Projects (IAF-ICP) (Award I2301E0026) administered by A*STAR, and by Alibaba Group and NTU Singapore through Alibaba-NTU Global e-Sustainability CorpLab (ANGEL).

\small
\bibliography{aaai25}

\newpage
\section{Supplementary Material} \label{supplementary-material}

In the supplementary material, we provide additional details on the experimental setup, experimental results, and visualizations to facilitate a deep understanding of our method for readers.

\subsection{A. Dataset Annotation}\label{app:dataset_annotation}
\begin{figure}[ht!]
	\centering
\includegraphics[width=0.99\linewidth]{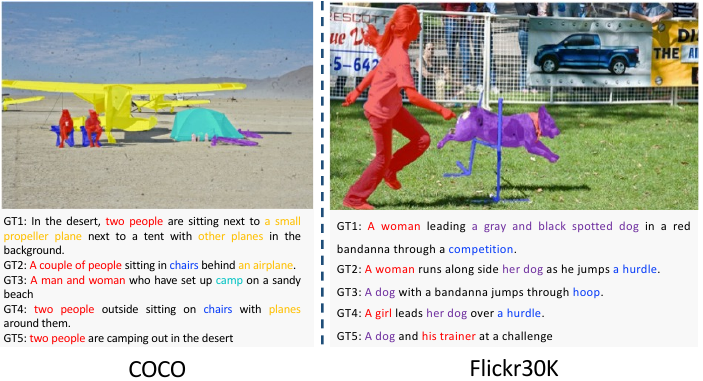}
	\caption{Two examples to illustrate our automatic annotation results on COCO and Flickr30k datasets.}
	\label{fig:auto_anno}
\end{figure}

Before starting our SegCaptioning task, it is essential to address the lack of ground truth for mask-word pairs in existing datasets. These pairs represent the grounding of each phrase in the captions corresponding to the matching object region in the images. In our experiments, we select COCO and Flickr30k-Entities datasets as our experimental data since they both contain captions and visual object annotations, which could satisfy the requirements of our task. We automatically annotate the mask-word pairs of the two datasets as follows: For the COCO dataset with labeled segmentation masks and captions, we directly match each noun in a caption with the 
text label of masks to build the mapping relationship between mask and caption words. For the Flickr30k-Entities dataset including the ground truth of captions and boxes for each image, we approximately transfer each box into a mask using the large-scale segmentation model SAM. Then, we establish the mapping relationship between masks and caption words following the way in the COCO dataset. Figure \ref{fig:auto_anno} shows two examples of the mask-word annotations from COCO and Flickr datasets, respectively.

\begin{figure*}[t!]
	\centering
\includegraphics[width=1\linewidth]{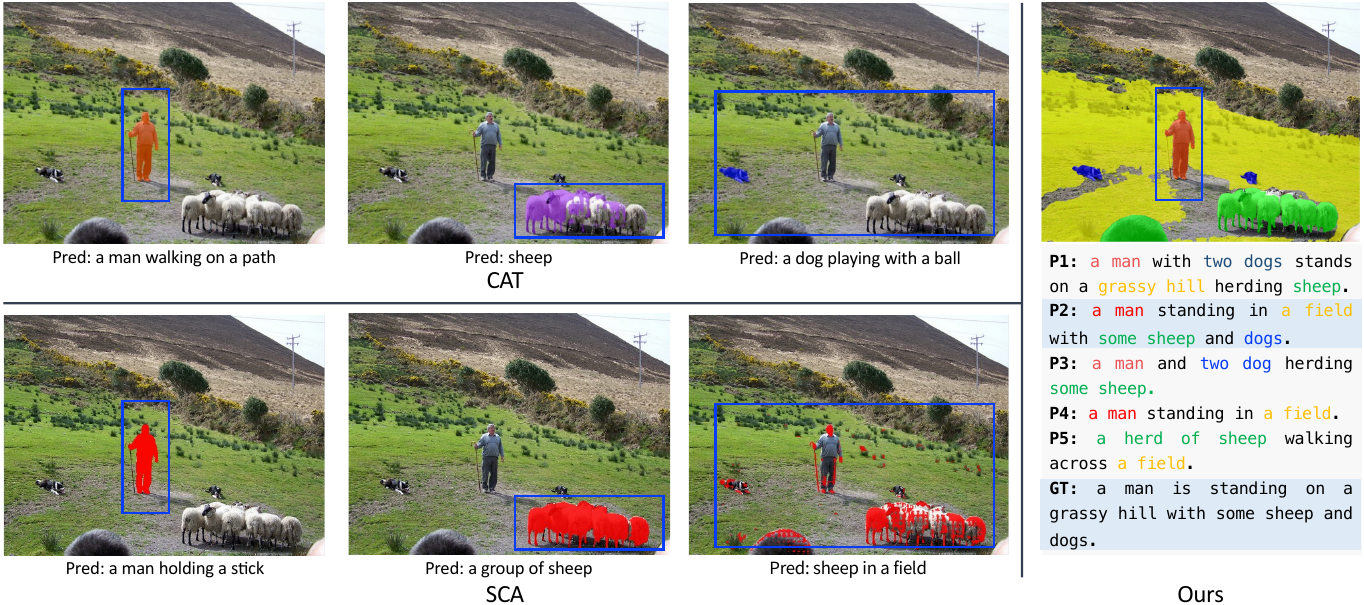}
	\caption{An example to visualize the segmentation and captioning results of our SGDiff, CAT and SCA conditioned on different box prompts.}
	\label{fig:multibox_case}
\end{figure*}

\begin{figure*}[t!]
	\centering
\includegraphics[width=1\linewidth]{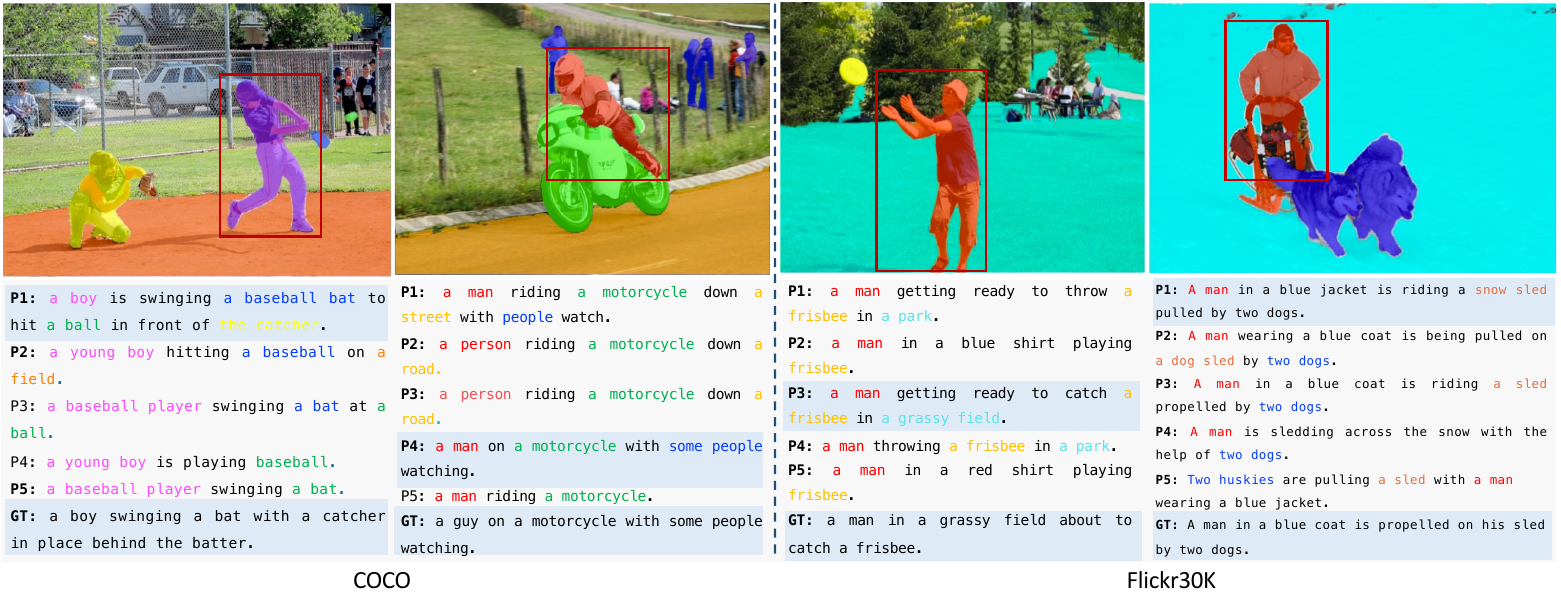}
	\caption{Additional visualization results of our method on the COCO and Flickr30k datasets.}
	\label{fig:more_results}
\end{figure*}

\subsection{B. Architecture and Training Details}\label{app:architecture_training}
Our PSGA generates subgraphs with a node size of $36$ and an edge size of $64$. We use a threshold $\theta = 0.5$ in Eq. (5) to filter out irrelevant nodes. During the denoising process with a timestep $T=50$, the generated captions have a maximum length of $20$ words. The SGBTrans consists of $6$ blocks that output features for captions and masks, and these features are then fed to the caption and mask generators, following the settings in \cite{luo2023semantic} and \cite{cheng2022masked}. We set ${\lambda}_1 = 2$ and ${\lambda}_2 = 1$ in Eq. (13). 
Our model training is divided into two phases. In the first phase, we train the captioning branch until the loss in the captioning branch converges, which ensures that the generated captions are coherent and closely aligned with the ground truth. Then, we proceed to the second phase to implement captioning-segmentation joint training. 
During the early stages of model convergence, since the quality of predicted masks is relatively poor, we use ground truth masks for alignment with captions when calculating the loss $\mathcal{L_{\text{MEC}}}$. After $10$ epochs, we switch to using the predicted masks for alignment.

\subsection{C. Comparative Analysis of Different Methods}\label{app:comparative_analysis}
We compare our method with CAT and SCA, which sequentially connect large vision models with large language models to output both masks and captions. Given the box prompt on an image (see the left column in Figure \ref{fig:multibox_case}), CAT and SCA only output a single mask for the box with a simplified caption to describe the object. Beyond this, our method could automatically predict related objects such as the dog and sheep using sophisticated captions to indicate their relationship. Furthermore, we tried to input a box containing multiple objects of the same type (see the second column in Figure \ref{fig:multibox_case}) to observe the segmentation and caption generation by CAT and SCA. We discover that CAT only generates a caption word ``sheep'' with inaccurate masks on the flock of sheep. SCA could successfully segment the flock, but the generated caption ``a group of sheep'' only focuses on the labeled box areas without deeply exploring its contextual information. 
Finally, we input a large bounding box encompassing multiple interested objects in the image (see the third column in Figure \ref{fig:multibox_case}). CAT and SCA not only fail to segment objects accurately, but the generated captions are also overly simplistic and incomplete. This result proves that they lack the capability of captioning and segmenting multiple objects. In contrast, our method could accurately capture the relationship between the flock and the people in nature language, incorporating exact segmentation results on all the related objects.

\subsection{D. Challenging Cases Visualization }\label{app:challenging_cases}
We present several challenging cases on both datasets to see the performance of our approach, as illustrated in Figure \ref{fig:more_results}. In the first and third examples, which feature complex backgrounds, our approach successfully segments related objects based on the scene relationships described in the captions, effectively excluding unrelated background elements such as crowds.
However, in the second example, a group of spectators is not fully segmented due to varying degrees of occlusion, resulting in some severely occluded individuals not being captured. 
Additionally, when it is necessary to simultaneously segment both the whole object and its parts, overlapping masks may occur, as seen in the fourth example with the caption ``A man in a blue coat''. We plan to address these limitations in future work.

\section{E. Inference Speed and Model Parameters}\label{app:speed_params}
Our model has 853 MB of parameters and performs single-image inference at 1.18 FPS, utilizing 14.3 GB of memory.
We compare the inference speed of our method with CAT, SCA, and GLaMM, which achieve 0.49 FPS, 0.65 FPS, and 0.14 FPS, respectively. 
Although these methods, which incorporate large language models, require fewer trainable parameters, their inference speed is significantly slower due to the computational overhead associated with large language models.
In contrast, our approach, utilizing a more efficient algorithmic framework, surpasses their performance and achieves faster inference speed.

\end{document}